\documentclass[preprint,twocolumn,5p,12pt]{article}

\usepackage{amssymb}
\usepackage{subcaption}
\usepackage{graphicx}
\usepackage{caption}
\usepackage{xspace}
\usepackage{multirow}
\usepackage{rotating}
 \usepackage{url}
\usepackage{gensymb}
\usepackage [ table ]{ xcolor }
\usepackage[top=2cm, bottom=2cm, left=1.5cm, right=1.5cm]{geometry}
\usepackage{setspace}
\usepackage{morefloats}

\usepackage{commath}
\usepackage[affil-it]{authblk}
\usepackage[colorlinks=true,citecolor=blue, urlcolor=blue, linkcolor=blue]{hyperref}
\usepackage{setspace}

\title{BS4NN: Binarized Spiking Neural Networks with Temporal Coding and Learning}%{Supervised spiking neural networks with one spike per neuron}

\author{Saeed Reza Kheradpisheh$ ^{1,}$\footnote{Corresponding Author \\Email addresses: \href{mailto://s_kheradpisheh@sbu.ac.ir}{s\_kheradpisheh@sbu.ac.ir} (SRK), \\ \href{mailto://mirsadeghi@aut.ac.ir}{mirsadeghi@aut.ac.ir} (MM), \\ \href{mailto:/timothee.masquelier@cnrs.fr}{timothee.masquelier@cnrs.fr} (TM)} }
\author{Maryam Mirsadeghi$ ^{2}$}
\author{Timoth\'ee Masquelier$ ^{3}$}
\affil{\footnotesize $ ^{1} $ Department of Computer Science, Faculty of Mathematical Sciences, Shahid Beheshti University, Tehran, Iran}
\affil{\footnotesize $ ^{2} $  Department of Electrical Engineering, Amirkabir University of Technology, Tehran, Iran}
\affil{\footnotesize $ ^{3} $  CerCo UMR 5549, CNRS Universit\'e Toulouse 3, France}

\date{}

\usepackage[absolute]{textpos}
\begin{document}
\begin{textblock}{19}(1,1)
\noindent \textbf{\color{red} This manuscript is published in \textbf{Neural processing Letters}. Please cite it as:}\\
\textit{\color{blue} S.R. Kheradpisheh, M. Mirsadeghi, T. Masquelier,  BS4NN: Binarized Spiking Neural Networks with Temporal Coding and Learning. Neural Processing Letters (2021).\\doi: \url{https://doi.org/10.1007/s11063-021-10680-x}}
\end{textblock}
\maketitle

\begin{abstract}

We recently proposed the S4NN algorithm, essentially an adaptation of backpropagation to multilayer spiking neural networks that use simple non-leaky integrate-and-fire neurons and a form of temporal coding known as time-to-first-spike coding. With this coding scheme, neurons fire at most once per stimulus, but the firing order carries information. Here, we introduce BS4NN, a modification of S4NN in which the synaptic weights are constrained to be binary (+1 or -1), in order to decrease memory (ideally, one bit per synapse) and computation footprints. This was done using two sets of weights: firstly, real-valued weights, updated by gradient descent, and used in the backward pass of backpropagation, and secondly, their signs, used in the forward pass. Similar strategies have been used to train (non-spiking) binarized neural networks. The main difference is that BS4NN operates in the time domain: spikes are propagated sequentially, and different neurons may reach their threshold at different times, which increases computational power. We validated BS4NN on two popular benchmarks, MNIST and Fashion-MNIST, and obtained reasonable accuracies for this sort of network (97.0\% and 87.3\% respectively) with a  negligible accuracy drop with respect to real-valued weights  (0.4\% and 0.7\%, respectively). We also demonstrated that BS4NN outperforms a simple BNN with the same architectures on those two datasets (by 0.2\% and 0.9\% respectively), presumably because it leverages the temporal dimension. The source codes of the proposed BS4NN are publicly available at \url{https://github.com/SRKH/BS4NN}.
\end{abstract}

\section{Introduction}\label{sec:introduction}
Spiking neural networks (SNNs), as the third generation of neural networks, are getting more and more attention due to their higher biological plausibility, hardware friendliness, lower energy demand, and temporal nature~\cite{tavanaei2018deep,Pfeiffer2018,taherkhani2020review,illing2019biologically}. Although SNNs have not yet reached the performance of the state-of-the-art artificial neural networks (ANNs) with deep architectures, recent efforts on adapting the gradient descent and  backpropagation algorithms to SNNs have led to great achivements~\cite{wang2020supervised}.

Contrary to artificial neurons with floating-point outputs, spiking neurons communicate via sparse and asynchronous stereotyped spikes which makes them suitable for event-based computations~\cite{tavanaei2018deep,Pfeiffer2018}. That is why the neuromorphic implementation of SNNs can be far less energy-hungry than ANN implementations~\cite{Roy2019},  which makes them appealing for real-time embedded AI systems and edge computing solutions. However, as SNNs become larger they require more storage and computational power. Binarizing the synaptic weights, similar to the binarized artificial neural networks (BANNs)~\cite{simons2019review}, could be a good solution to reduce the memory and computational requirements of SNNs.

Although the use of  binary (+1 and -1) weights in ANNs is not a very recent idea~\cite{saad1990training,venkatesh1993directed,baldassi2007efficient}, the early studies could not adapt backpropagation to BANNs. Since binary weights cannot be updated in small amounts, the backpropagation and stochastic gradient descent algorithms cannot be directly applied to BANNs. By proposing BinaryConnect~\cite{courbariaux2015binaryconnect,courbariaux2016binarized} Courbariaux et al. were the first who successfully trained deep BANNs using the backpropagation algorithm. They used real-valued weights which are binarized before being used in the forward pass. During backpropagation, using the Straight-Through Estimator (STE),  the gradients of the binary weights are simply passed and applied to the real-valued weights.  Soon after, Rastegari et al.~\cite{rastegari2016xnor} proposed XNOR-Net that is very similar to BinaryConnect but it multiplies a per-layer scaling factor (the L1-norm of real-valued weights) to the binary weights to make a better approximation of the real-valued weights. In order to speed up the learning phase of BANNs, Tang et al.~\cite{tang2017train}  controlled the rate of oscillations in binary weights between -1 and 1 by optimizing the learning rates. They also proposed to use learned scaling factors instead of the L1-norm of real-valued weights in XNOR-Net. In DoReFa-NET~\cite{zhou2016dorefa}, Zhou et al. proposed a model with variable width-size (down to binary) weights, activations, and even gradients during backpropagation. A more detailed survey on BANNs is provided in~\cite{simons2019review}.

%stochastic STDP for binary weights
A few recent studies have tried to convert supervised BANNs into equivalent binary SNNs (BSNNs), however, there is no other study to the best of our knowledge aimed at directly training multi-layer supervised SNNs with binary weights. Esser et al.~\cite{NIPS2015_5862} trained ANNs with constrained weights and activations and deployed them into SNNs with binary weights on TrueNorth.  Later in~\cite{Esser11441}, they mapped convolutional ANNs with trinary weights and binary activations to SNNs on TrueNorth. Rueckauer et al.~\cite{rueckauer2017conversion} converted  BinaryConnect~\cite{courbariaux2015binaryconnect} with binary and full-precision activations into  equivalent rate-coded BSNNs. Although their converted BSNN had binary weights, they did not binarize the full-precision parameters of the batch-normalization layers.  In~\cite{wang2020deep}, Wang et al. convert BinaryConnect networks to rate-coded BSNNs using a weights-thresholds balance conversion method which scales the high-precision batch normalization parameters of BinaryConnect into -1 or 1.  In another study, Lu et al.~\cite{lu2020exploring} converted a modified version of XNOR-Net without batch normalization and bias inputs into equivalent rate-coded BSNNs. 

In this work, we propose a direct supervised learning algorithm to train multi-layer SNNs with binary synaptic weights. The input layer uses a temporal time-to-first-spike coding~\cite{kheradpisheh2018stdp,mozafari2018first,Kheradpisheh2015} to convert the input image into a spike train with one spike per neuron. The non-leaky integrate-and-fire (IF) neurons in the subsequent hidden and output layers integrate incoming spikes through binary (+1 or -1) synapses and emit only one spike right after the first crossing of the threshold. Inspired by BANNs, we also use a set of real-valued proxy weights such that the binary weights are indeed the sign of real-valued weights.  Hence, in the backward pass, we update the real-valued weights based on the errors made by the binary weights. Literally, after completing the forward pass with binary weights, the output layer computes the errors by comparing its actual and target firing times, and then, real-valued synaptic weights get updated using the temporal error backpropagation. We evaluated the proposed network on MNIST~\cite{lecun1998gradient} and Fashion-MNIST~\cite{xiao2017fashion} datasets with 97.0\% and 87.3\% categorization accuracies, respectively.
 
SNNs can vary in terms of neuronal model, neural connectivity, information coding, and learning strategy, which deeply affect their accuracy, memory, and energy efficiency. The advantages of the proposed BSNN are 1) the use of non-leaky IF neurons with a very simple neuronal dynamics, 2) having binarized connectivity with low memory and computational cost 3) the use of a sparse temporal coding with at most one spike per neuron, and 4) learning by a direct supervised
temporal learning rule which forces the network to make decisions as accurate and early as possible.

\section{Binarized Spiking Neural Networks}
The input layer of the  proposed  \textit{binarized single-spike supervised spiking neural network} (BS4NN) converts the input image into a spike train based on a time-to-first-spike coding. These spikes are then propagated through the network, where, the binary IF neurons in hidden and output layers are not allowed to fire more than once per image. Each output neuron is dedicated to a different category and the first output neuron to fire determines the decision of the network. 

The error of each output neuron is computed by comparing its actual firing time with a target firing time. Then, a modified version of the temporal  backpropagation algorithm in S4NN~\cite{Kheradpisheh2020} is used to update the synaptic weights. During the learning phase, we have two sets of weights, the real-valued weights, $W$, and the corresponding binary weights, $B$, where $B=sign(W)$. The  forward propagation is done with the binary weights, while, the error backpropagation and weight updates are done by the real-valued weights. Finally, we put the real-valued weights aside and use the binary weights to inference about testing images. Note that some of the following equations are adopted from S4NN~\cite{Kheradpisheh2020} and they are reproduced here for the sake of readers.

\subsection{Forward pass}
The input layer  converts the input image into a volley of spikes using a single-spike temporal coding scheme known as intensity-to-latency conversion. For images with the pixel intensity of range $[0,I_{max}]$, the firing time of the $i$th input neuron, $t_i$, corresponding to the $i$th pixel intensity, $I_i$, is computed as
 \begin{equation}
     t_i = \left\lfloor \frac{I_{max}-I_i}{I_{max}} \: t_{max}\right\rfloor,
 \end{equation}
where, $t_{max}$ is the maximum firing time. In this way, input neurons with higher pixel intensities have shorter spike latencies. Here, we used discrete time. Therefore, the spike pattern of the $i$th input neuron is defined as
\begin{equation}
     S_i^0(t)=
     \begin{cases}
     1 & \quad \text{if  } t=t_i\\
     0 & \quad \mathrm{otherwise}.
     \end{cases}
 \end{equation}

Subsequent hidden and output layers are comprised of non-leaky IF neurons. The $j$th IF neuron of $l$th layer receives incoming spikes through binary synaptic weights of -1 or +1 and update its membrane potential, $V_j^l$, as

 \begin{equation}\label{Eq:IFNeuronModel}
     V_j^l(t)= V_j^l(t-1)+ \alpha^l\sum_i B_{ji}^{l} S_i^{l-1}(t),
 \end{equation}
 where $S_i^{l-1}$ and $B_{ji}^{l}$ are, respectively, the input spike pattern and the binary synaptic weight connecting the $i$th presynaptic neuron to the neuron $j$.  Note that $\alpha^l$ is a scaling factor shared between all the neurons of the $l$th layer. These scaling factors help the network to avoid the emergent of silent neurons (i. e., unable to reach their threshold for any stimuli) that is more probable with binary weights.

  The IF neuron fires only once, the first time its membrane potential crosses the threshold $\theta$,
 
  \begin{equation}\label{EQ}
     S_j^l(t)=
     \begin{cases}
     1 & \quad \text{if  } V_j^l(t) \geq \theta\:\&\: S_j^l(<t)\neq 1\\
     0 & \quad \mathrm{otherwise}.
     \end{cases}
 \end{equation}
 where $S_j^l(<t)\neq 1$ checks if the neuron has not fired at any  previous time step.  Equivalently, one can move the scaling factor $\alpha^l$ from Eq.~\ref{Eq:IFNeuronModel} to Eq.~\ref{EQ} by replacing $\theta$ with $\theta/\alpha^l$.
 
 For each input image, we first reset all the membrane voltages to zero and then run the simulation for at most $t_{max}$ time steps. Each output neuron is assigned to a different category and the output neuron that fires earlier than others determines the category of the input image. Hence, in the test phase, we do not need to continue the simulation after the first spike in the output layer. If none of the output neurons fires before $t_{max}$, the output neuron with the maximum membrane potential at $t_{max}$ makes the decision. However, during the learning phase, to compute the temporal error and gradients, we need all the neurons in the network to fire at some point, and hence, we continue the simulation until $t_{max}$ and if a neuron never fires, we force it to emit a fake spike at time $t_{max}$.

\subsection{Backward pass}
For a categorization task with $C$ categories, we define the temporal error as a function of the actual and target firing times, 
\begin{equation}\label{Eq:error}
    e=[e_1,...,e_C] \quad  \text{s.t.} \quad
    e_j=(T_j^o-t_j^o)/t_{max},
\end{equation}
where $t_j^o$ and $T_j^o$ are the actual and the target firing times of the $j$th output neuron, respectively. Define $\tau$ as the minimum firing time in the output layer (i. e., $\tau=min\lbrace t_j^o|1 \leq j \leq C\rbrace$). For an input image belonging to the $i$th category, we have
\begin{equation}
 T_j^o =
   \begin{cases}
   \tau-\gamma       & \quad \text{if } j =i,\\
     \tau+\gamma & \quad \text{if } j\neq i \: \: \& \: \: t_j^o<\tau+\gamma,\\
     t_j^o & \quad \text{if } j\neq i \:\: \& \:\: t_j^o\geq\tau+\gamma,
   \end{cases}
 \end{equation}
 where, $\gamma$ is a positive constant. This way the correct neuron is encouraged to fire first and others are penalized to not fire earlier than $\tau+\gamma$. In a special case that all the output neurons  remain silent during the forward pass  (emit fake spikes at $t_{max}$), we set $T_i^o =t_{max}-\gamma$  and $T_{j\neq i}^o =t_{max}$  to force the correct neuron to fire.  These target firing times enforce the network to respond as accurate and early as possible.
  
We define the ``mean squared error" (MSE) loss function as
\begin{equation}\label{EQ:loss}
   L=\frac{1}{2}\Vert e \Vert^2 =\frac{1}{2}\sum\limits_{j=1}^{C} e_j^2.
\end{equation}

To apply the gradient descent algorithm, we should compute $\partial L/\partial B_{ji}^l$, the gradient of the loss function with respect to the binary weights. However, the gradient descent method makes small changes to the weights, which cannot be done with binary values. To solve the problem, during the learning phase, we use a set of real-valued weights, $W$, as a proxy, such that  
\begin{equation}
B_{ji}^l=sign(W_{ji}^l),
\end{equation}
and, as the gradient of the $sign$ function is 0 or undefined, using the straight-through estimator (STE)   we approximate $\partial sign(x)/\partial x=1$, therefore, we have
\begin{equation}
   \frac{\partial L}{\partial W_{ji}^l}=\frac{\partial L}{\partial B_{ji}^l}.
\end{equation}

Now, we can update the real-valued weights as
 \begin{equation}\label{Eq:LearningRule}
   W_{ji}^l=W_{ji}^l-\eta \frac{\partial L}{\partial B_{ji}^l},
 \end{equation}
where $\eta$ is the learning rate parameter. 

%%%%%%%%%%%%%% result tables %%%%%%%%%%%%%%%%
 \begin{table*}
 \begin{center}
 \caption{The structural, initialization, and model parameters used for  MNIST and Fashion-MNIST datasets.}\label{table:parameters}

\resizebox{\textwidth}{!}{ \begin{tabular}{lccccccccccccccc}
 & \multicolumn{2} {c}{Layer size} && \multicolumn{2} {c}{Initial real-value weights}&&\multicolumn{2} {c}{ Learnable parameters}&&\multicolumn{6} {c}{ Hyper-parameters}\\ \cline{2-3}\cline{5-6}\cline{8-9} \cline{11-16}
  Dataset &  Hidden  &Output && $W^h$ & $W^o$ &&$\alpha^h$&$\alpha^o$&&$t_{max}$ & $\theta $ & $\eta$& $\mu$ &$\gamma$ &$\lambda$\\
 \hline
 MNIST &600&10&&$[0,5]$&$[0,50]$&& 5&5 &&256&100&0.1&0.01&1&$10^{-6}$\\
 Fashion-MNIST&1000&10&&$[0,1]$&$[0,1]$&&5&10 &&256&700&0.1&0.01&1&$10^{-6}$\\
 \end{tabular}}
 \end{center}
 \end{table*}
 
 \begin{table*}
 \begin{center}
 \caption{The recognition accuracies of recent supervised fully connected SNNs with spike-time-based backpropagation on the MNIST dataset. The details of each model including its input coding scheme, neuron model, synapses, post-synaptic potential (PSP), learning method, and the number of hidden neurons are provided.}\label{tbl1}
 \scriptsize
 \begin{tabular}{llllcc}
 Model              &Coding&Neuron / Synapse /  PSP & Learning&Hidden(\#) & Acc. (\%) \\
 \hline
 Mostafa (2017)~\cite{mostafa2017supervised} &Temporal&IF / Real-value /Exponential  & Temporal Backpropagation& 800&97.2   \\
 Tavanaei et al. (2019) ~\cite{tavanaei2019bp}& Rate &IF / Real-value  / Instantaneous&STDP-based
Backpropagation&1000& 96.6    \\
 Comsa et al. (2019)~\cite{comsa2019temporal} &Temporal& SRM / Real-value / Exponential&Temporal Backpropagation&340& 97.9  \\
  Zhang et al. (2020)~\cite{zhang2020spike}   & Temporal & IF / Real-value / Linear& Temporal Backpropagation&400&98.1   \\
  Zhang et al.(2020) ~\cite{zhang2020spike}   & Temporal & IF / Real-value / Linear& Temporal Backpropagation&800&98.4   \\
 Sakemi et al.(2020) ~\cite{sakemi2020supervised}   & Temporal & IF / Real-value / Linear& Temporal Backpropagation&500&97.8   \\
Sakemi et al.(2020) ~\cite{sakemi2020supervised}   & Temporal & IF / Real-value / Linear& Temporal Backpropagation&800&98.0   \\
S4NN~\cite{Kheradpisheh2020}   & Temporal & IF  / Real-value / Instantaneous &Temporal Backpropagation& 600&97.4   \\

BNN&Binary (0 \& 1)& Binary Sigmoid/ Binary/ -&Backpropagation with ADAM&600&96.8\\
 BS4NN (this paper)    & Temporal & IF / Binary / Instantaneous& Temporal Backpropagation&600&97.0   \\
 
 \end{tabular}
 \end{center}
 \end{table*}

%%%%%%%%%%%%%%%%%%%%%%%%%%%%%%%%%%%%%%%%%

We define
\begin{equation}\label{Eq:delta} 
     \delta_{j}^{l}=\frac{\partial L}{\partial t_j^l},
 \end{equation}
 where, $t_j^l$ is the firing time of the $j$th neuron of the $l$th layer. Also, according to~\cite{Kheradpisheh2020}, we approximate $\partial t_j^l/\partial B_{ji}^{l}$  to be $-\alpha^l$ if $t_i^{l-1}<t_j^l$ and 0 otherwise. Therefore,  we have
 \begin{equation}
     \frac{\partial L}{\partial B_{ji}^l}=\frac{\partial L}{\partial t_j^l}\frac{\partial t_j^l}{\partial B_{ji}^{l}}= \begin{cases}
     - \alpha^l \delta_j^l    &  \text{if } t_i^{l-1}<t_j^l\\
     0  &  \mathrm{otherwise},
   \end{cases}
 \end{equation}
where for the output layer (i. e., $l=o$) we have
\begin{equation}
     \delta_j^o=\frac{\partial L}{\partial e_j} \frac{\partial  e_j}{\partial t_j^o}=-\frac{e_j}{t_{max}},
 \end{equation}
 and for the hidden layers (i. e., $l \neq o$), according to the backpropagation algorithm, we compute  the weighted sum of the delta values of neurons in the following layer,
  \begin{equation}\label{hhhhhhh}
      \begin{split}
      \delta_j^l &=
      \sum_{k}\frac{\partial L}{\partial t_{k}^{l+1}}\frac{\partial t_{k}^{l+1}}{\partial V_{k}^{l+1}}\frac{\partial V_{k}^{l+1}}{\partial t_{j}^{l}} \\
      & =\sum_{k}\delta_k^{l+1}\alpha^{l+1}B_{kj}^{l+1}[t_j^l < t_k^{l+1}],
      \end{split}
  \end{equation}
 where, $k$ iterates over neurons in layer $l+1$. Similar to~\cite{Kheradpisheh2020},  we approximate $\partial t_{k}^{l+1}/\partial V_{k}^{l+1}=-1$ and $\partial V_{k}^{l+1}/\partial t_{j}^{l}= -\alpha^{l+1} B_{kj}^{l+1}$ if and only if $[t_j^l \leq t_k^{l+1}]$. To have smooth gradients, we use the real-valued weights, $W_{kj}^{l+1}$, instead of  the  scaled binary weights, $\alpha^{l+1} B_{kj}^{l+1}$.
 
 We also update the scaling factor $\alpha^l$ as 
 \begin{equation}
 \alpha^l=\alpha^l-\mu\frac{\partial L}{\partial \alpha^l}
 \end{equation}
 where $\mu$ is the learning rate parameter. Therefore we compute
 
  \begin{equation}\label{alphaeq}
  \begin{split}
  \frac{\partial L}{\partial \alpha^l} &=  \sum_j \frac{\partial L}{\partial t_j^l}\frac{\partial t_j^l}{\partial \alpha^{l}}  =\sum_j    \frac{\partial L}{\partial t_j^l} \frac{\partial t_j^l}{\partial V_j^{l}}\frac {\partial V_j^{l}}{\partial \alpha^{l}} \\
  &=-\sum_j \delta^l_i \sum_i B_{ji}^l[t_i^{l-1}<t_j^l]
  \end{split}
  \end{equation}
 where $j$ and $i$ iterate over neurons in layer $l$ and $l-1$, respectively. Here again, similar to~\cite{Kheradpisheh2020}, we approximate 
$\partial t_j^l/\partial V_j^{l}=-1$ and according to Eq~.\ref{Eq:IFNeuronModel}, we compute $\partial V_j^{l}/\partial \alpha^{l} = \sum_i B_{ji}^l[t_i^{l-1}<t_j^l]$.  

Note that one of the main challenges with single-spike coding schemes in SNNs is the emergent of dead neurons during training. Technically, when, during the learning of a specific stimulus, you decrease a neuron's weights to enforce it to fire later, there is a chance that the weights change in an amount that the neuron is no more able to reach its threshold even for other input stimuli. This situation is more extreme in binary weights because the weigh changes are not graded and the weights can suddenly change from 1 to -1. To avoid such a situation, we added the layer-wised weight scaling factor  $\alpha^l$ to our neuron model and according to Eq.~\ref{alphaeq} this factor can change smoothly and prevent such a drastic situation. 

Before updating the weights we normalize the gradients as  $\delta_j^l= \delta_j^l/\sum_i \delta_i^l$, to avoid  exploding and vanishing gradients. Also, we added an $L_2$-norm weight regularization term $\lambda \sum_{l}\sum_{i,j}(W_{ji}^l)^2$ to the loss function to avoid overfitting. The parameter $\lambda$ is the regularization parameter accounting for the degree of weigh penalization.

\section{Results}
\subsection{MNIST dataset}
In this section, we evaluate BS4NN on the MNIST  dataset, which is the most popular benchmark for spiking neural networks~\cite{tavanaei2018deep}. The MNIST dataset contains 60,000 handwritten digits (0 to 9) in images of size $28\times28$ pixels as the train set. The test set contains 10,000 digit images, $\sim1000$ images per digit. Here, we train a fully connected BS4NN with one hidden layer containing 600 IF neurons. The parameter settings are provided in Table~\ref{table:parameters}. Initial synaptic weights including input-hidden  ($W^h$ ) and hidden-output ($W^o$) weights are drawn from uniform distributions in range $[0,5]$ and $[0,50]$, respectively. Trainable parameters including the synaptic weights the scale factors of hidden ($\alpha^h$) and output ($\alpha^o$) layers are tuned through the learning phase. Adaptive parameters including $\eta$ and $\mu$ are decreased by 30\% every ten epochs. Other parameters remain intact in both the learning and testing phases.

\begin{figure}
     \centering
     \includegraphics[width=0.5\textwidth]{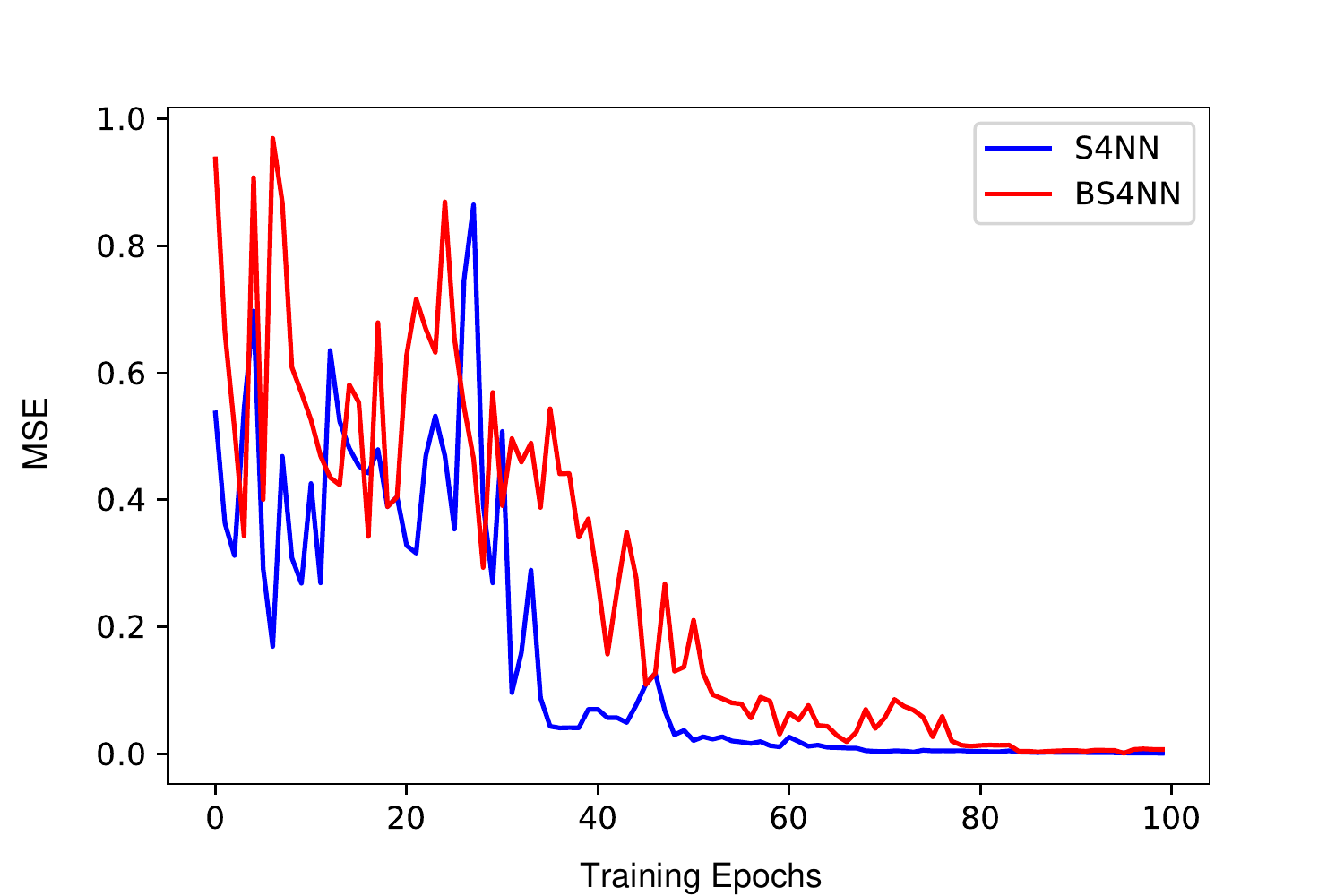}
     \caption{The MSE learning curves of  BS4NN and S4NN models on the training samples of the MNIST dataset during the training phase.}
     \label{MSE}
 \end{figure}

Table~\ref{tbl1} presents the categorization accuracy of the proposed BS4NN along with some other SNNs with spike-time-based direct supervised learning algorithms and fully-connected architectures. BS4NN is the only network in this table that uses binary weights and it could reach 97.0\% accuracy on MNIST. As mentioned in the Methods Section, BS4NN uses a modified version of the temporal backpropagation algorithm in S4NN (Kheradpisheh et al. (2020)~\cite{Kheradpisheh2020}) to have binary weights. Compared to S4NN, the categorization accuracy in BS4NN dropped by 0.4\% only. Although BS4NN is outperformed by the other SNNs by at most 1.4\%, its advantages are the use of binary weights instead of real-valued full-precision weights and instantaneous post-synaptic potential (PSP) function. As seen, BS4NN could outperform Tavanaei et al. (2019) ~\cite{tavanaei2019bp} that uses real-valued weights and instantaneous PSP. Other SNNs use exponential and linear PSP functions that complicate the neural processing and the learning procedure of the network, which consequently, increases their computational and energy cost. The MSE learning curves of BS4NN and S4NN models during the training phase and over the training samples are illustrated in Figure~\ref{MSE}. The MSE curve of BS4NN fluctuates to a higher extent than that of  S4NN which is due to the sudden flip of the binary weights of BS4NN, however, it finally relaxes down and converges to small values around zero.

   \begin{figure}[!htb]
     \centering
     \begin{subfigure}[b]{0.5\textwidth}
     \includegraphics[width=\textwidth]{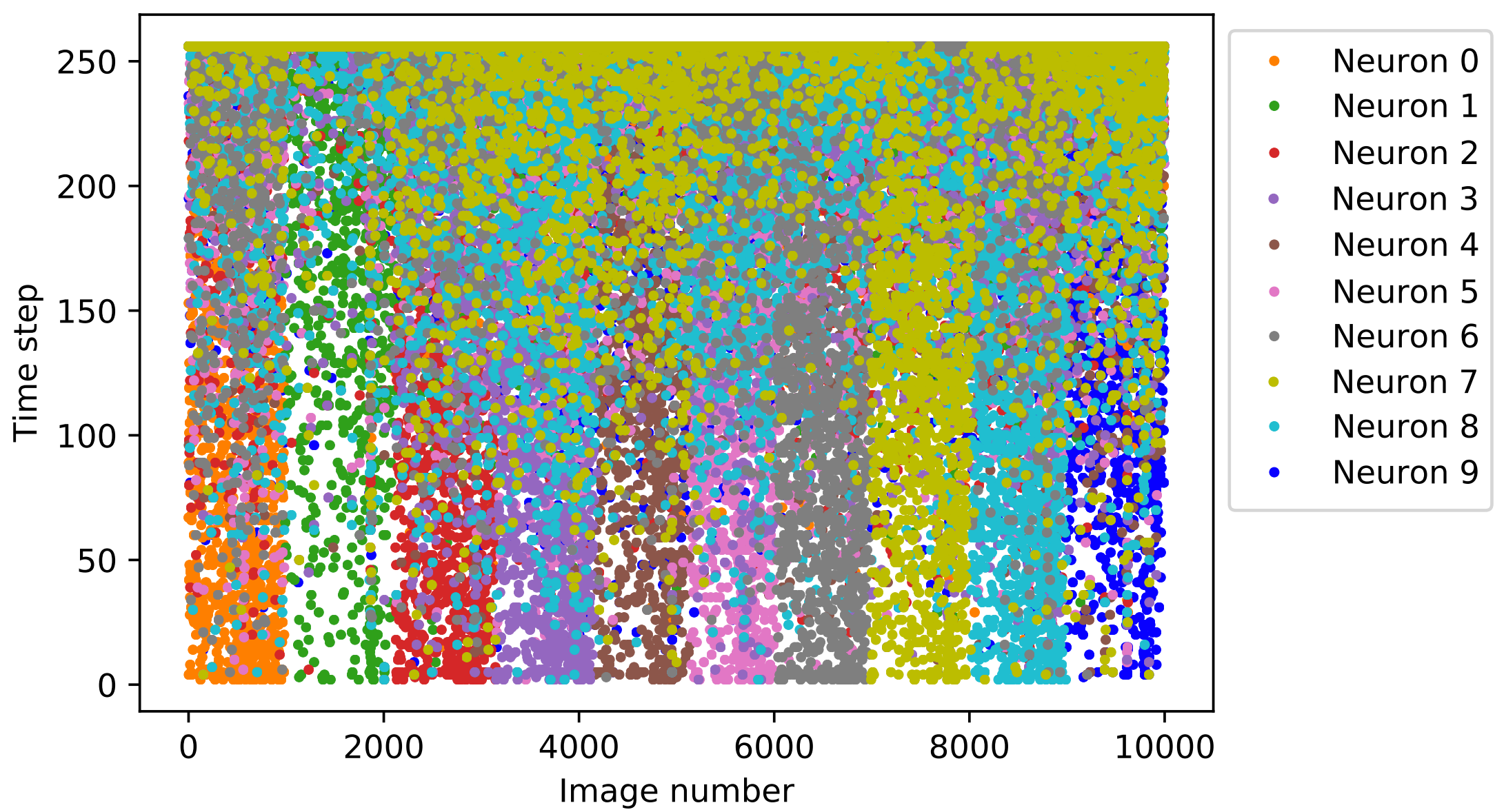}
        \caption{}
        \label{fig:1A}
    \end{subfigure} 
          \begin{subfigure}[b]{0.4\textwidth}
     \includegraphics[width=\textwidth]{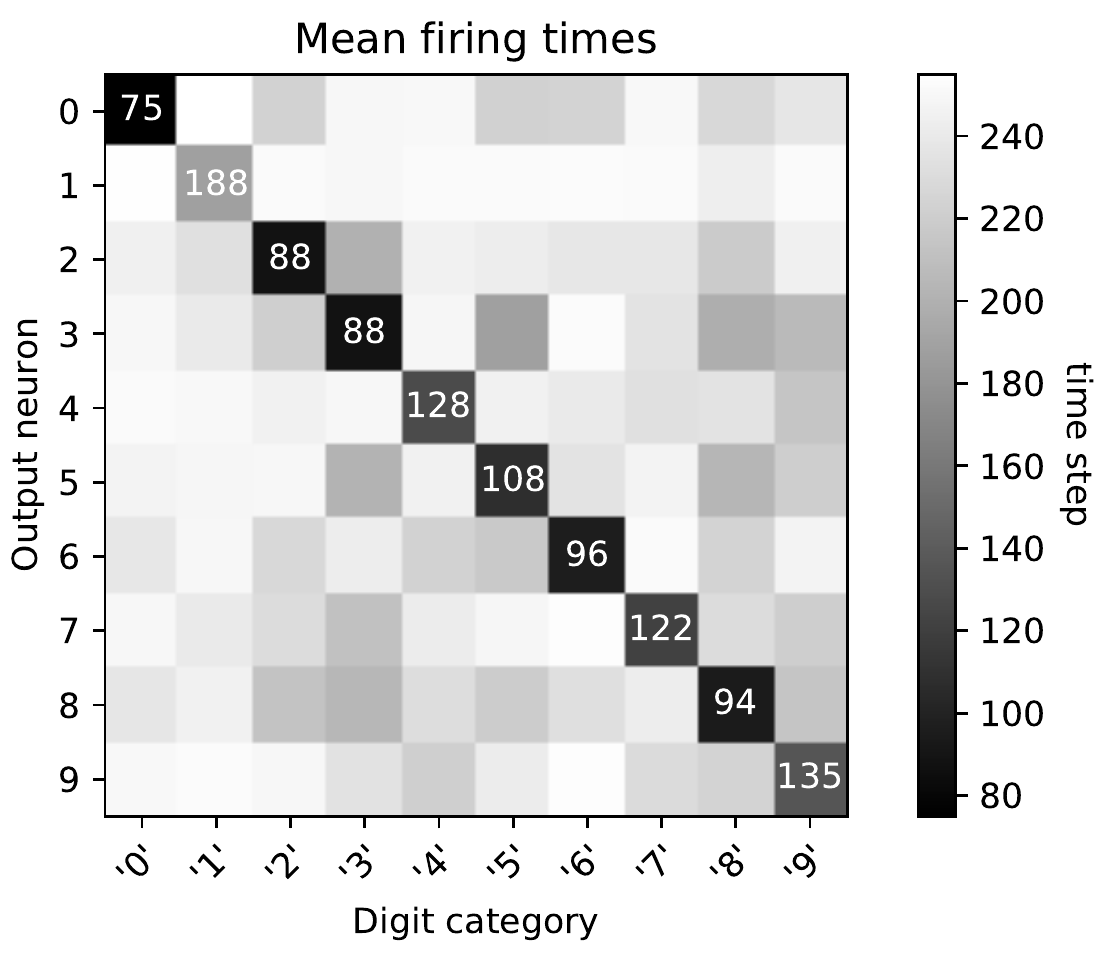}
        \caption{}
        \label{fig:1B}
    \end{subfigure}
\caption{(a) The firing times of the ten output neurons over the test images ordered by category. (b) The mean firing time of each output neuron (rows) over the images of different digits (columns).}\label{fig:1}
 \end{figure}

 %%%%%%%%% BNN %%%%%%%%%%%
 We also compared BS4NN to a BNN with a similar architecture.  To do a fair comparison, inspired from~\cite{courbariaux2016binarized}, we implemented a BNN with binary weights (-1 and +1)  and binary sigmoid activations (0 and 1). The network has a single hidden layer of size 600 and it is trained using ADAM optimizer and squared hinge loss function for 500 epochs.  The learning rate initiates from $10^{-3}$ and exponentially decays, through the learning epochs, down to  $10^{-4}$. According to~\cite{glorot2010understanding}, we initialize the real-valued weights of each layer with random values drawn from a uniform distribution in range  $[-1/\sqrt{n}, 1/\sqrt{n}]$, where, $n$ is the number of synaptic weights of that layer. As provided in Table~\ref{tbl1}, the BNN could reach the best accuracy of 96.8\% on MNIST, which is a 0.2\% drop with respect to BS4NN (we will comment on these results in the Discussion).
 
  \begin{figure}
     \centering
     \includegraphics[width=0.4\textwidth]{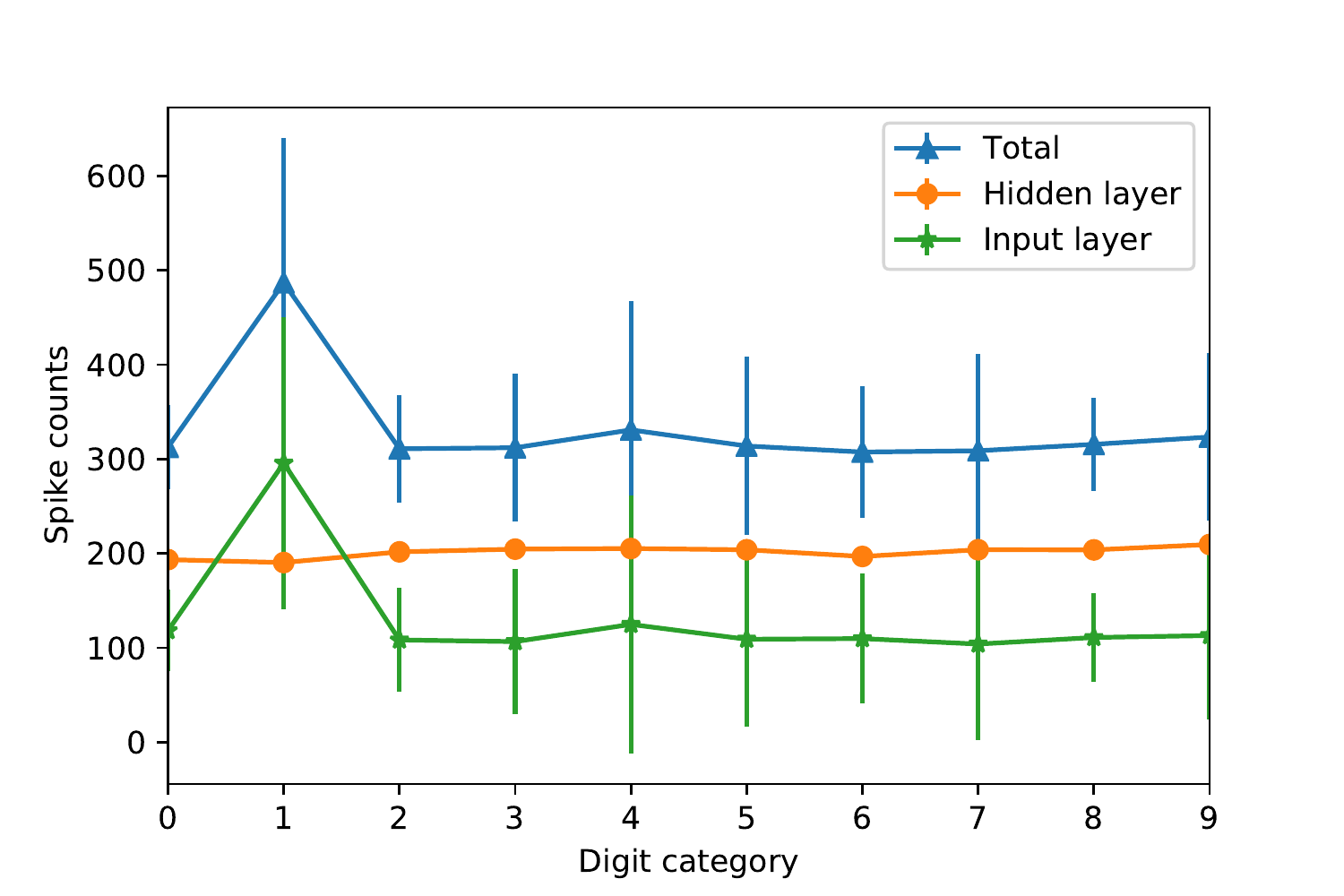}
     \caption{The mean required number of spikes in the input, hidden, and total layers.}
     \label{fig:2}
 \end{figure} 

The firing times of the ten output neurons over all test images are shown in Figure~\ref{fig:1A}. Images are ordered by the digit category from '0' to '9'. For each test image, the firing time of each neuron is shown by a color-coded dot. As seen, for each category, its corresponding output neuron tends to fire earlier than others. It is better evident in Figure~\ref{fig:1B} which shows the mean firing time of each output neuron for each digit category. Each output neuron has, by difference, the shortest mean firing time for images of its corresponding digit. Interestingly, BS4NN needs a much longer time to detect digit '1' (188 time steps) that could be due to the use of binary weights. Other digits cover more pixels of the image, and therefore, produce more early spikes than digit '1'. Since the weights are binary, the few early spikes of digit '1' can not activate the hidden IF neurons, and hence, BS4NN needs to wait for later surrounding spikes to distinguish digit '1' from other digits. 

 We further counted the mean required number of spikes for BS4NN to categorize images of each digit category. To this end, we counted the number of spikes in all the layers until the emission of the first spike in the output layer (when the network makes its decision). The mean required spikes of the input and hidden layers are depicted in Figure~\ref{fig:2}. All  digit categories but '1', on average, require about 100 spikes in the input and 200 spikes in the hidden layers, respectively.  Digit '1' requires about 300 input spikes, while, similar to other digits, its hidden layer needs about 100 spikes. As explained above, digit '1' covers a fewer number of pixels than other digits and also its shape overlaps with the constituent parts of some other digits, hence, due to the use of the binary weights, the network should wait for later input spikes to distinguish digit '1' from other digits.

\begin{figure}
     \centering
     \includegraphics[width=0.5\textwidth]{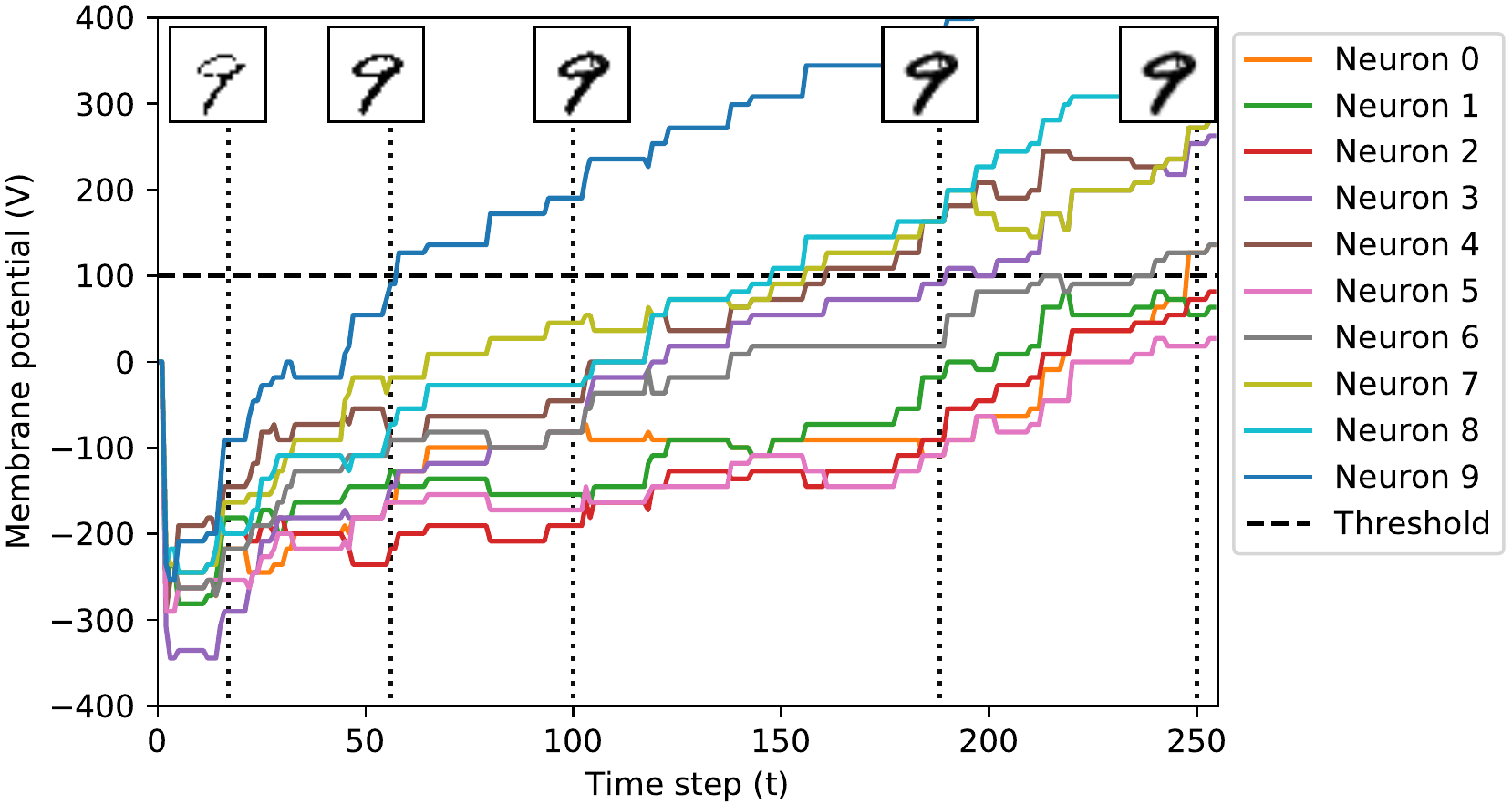}
     \caption{The trajectory of the membrane potential for all the ten output neuron for sample '9' test image along with the demonstration of the  accumulated input spikes until the 15, 58, 100, 190, and 250 time steps}
     \label{fig:3}
 \end{figure}

Figure~\ref{fig:3} shows the time course of the membrane potentials of the output neurons for a sample '9' test image. The membrane potential of the 9th output neuron overtakes others at the 15th time step and quickly increases until it crosses the threshold at the 58th time step. 
The accumulated input spikes until the 15, 58, 100, 190, and 250 time steps are depicted in this figure. As seen, up to the 15th time step, a few input spikes are propagated and at the 58th time steps with the propagation of a few more input spikes, the 9th output neuron reaches its threshold and determines the category of the input image. Later input, hidden, and output spikes are no more required by the network.

\begin{figure}
     \centering
     \begin{subfigure}[b]{0.5\textwidth}
     \includegraphics[width=\textwidth]{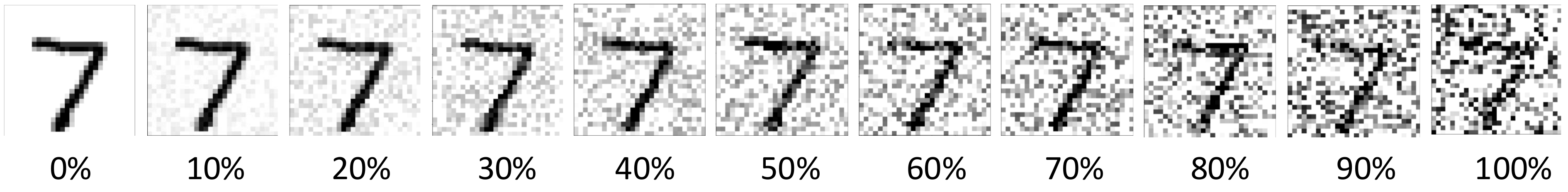}
        \caption{}
        \label{fig:4A}
    \end{subfigure} 
          \begin{subfigure}[b]{0.4\textwidth}
     \includegraphics[width=\textwidth]{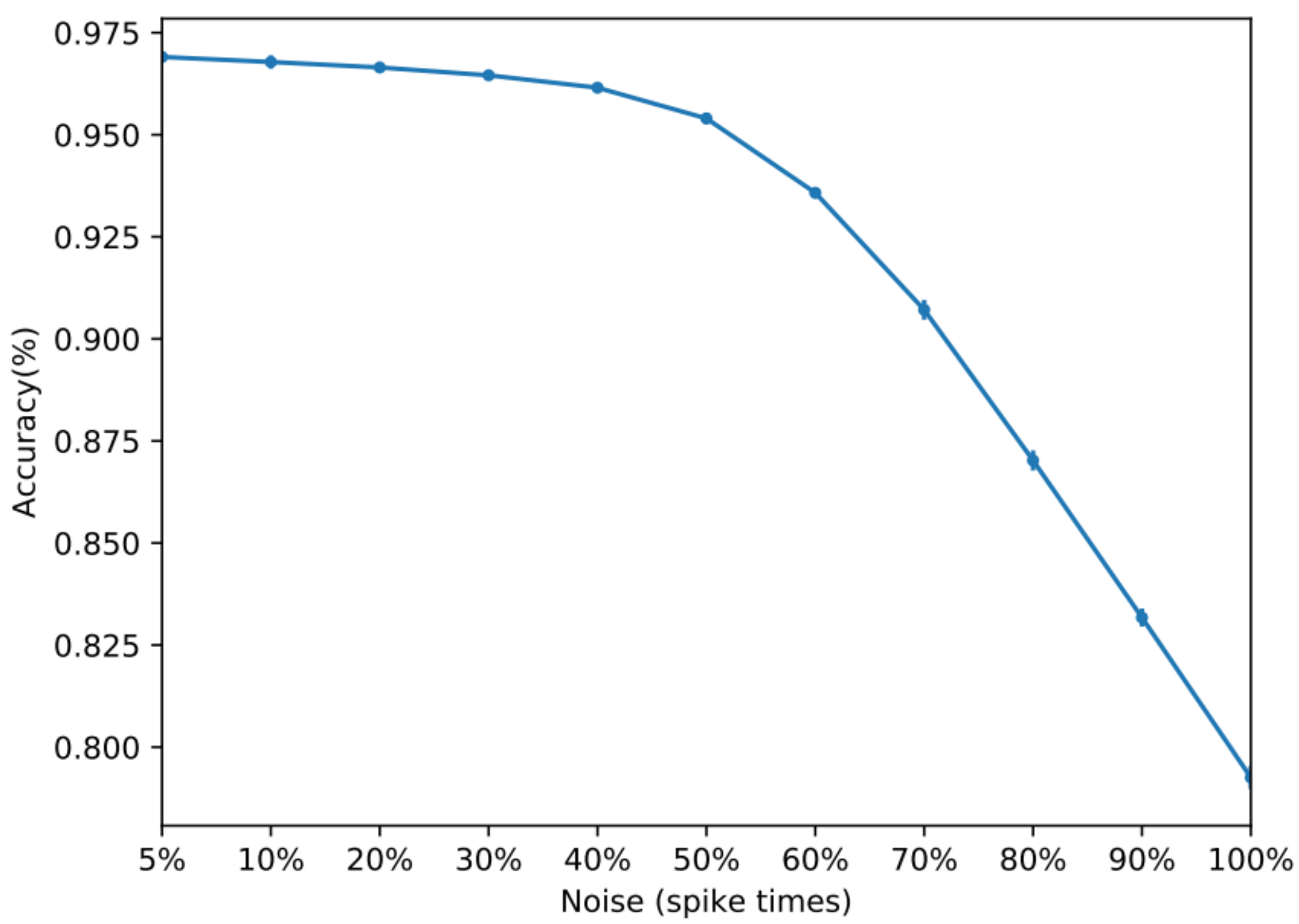}
        \caption{}
        \label{fig:4B}
    \end{subfigure}
\caption{(a) A sample image contaminated with different amount of jitter noise. (b) The recognition accuracy of the trained BS4NN on test images under different levels of noise.}\label{fig:4}
 \end{figure}

To evaluate the robustness of the  trained BS4NN to the input noise, during the test phase, we added random jitter noise drawn from a uniform distribution in the range $[-J,J]$ to the pixels of the input images. The noise level, $J$, varies from 5\% to 100\% of the maximum pixel intensity, $I_{max}$. Figure~\ref{fig:4A} shows a sample image contaminated with different levels of jitter noise. The recognition accuracy of the trained model over noisy test images under different levels of noise is plotted in Figure~\ref{fig:4B}.  As shown, the recognition accuracy remains above 95\% and it drops to 79\% for the 100\% noise level, while, in S4NN~\cite{Kheradpisheh2020} with real-valued weights, the accuracy drops at most 5\% for the 100\% noise level. In higher noise levels, the order of input spikes can dramatically change and because BS4NN has only +1 and -1 synaptic weights even to the insignificant parts of the input images, It affects the behavior of IF neurons which consequently increases the categorization error rate.
 
  \begin{figure}
     \centering
     \includegraphics[width=0.5\textwidth]{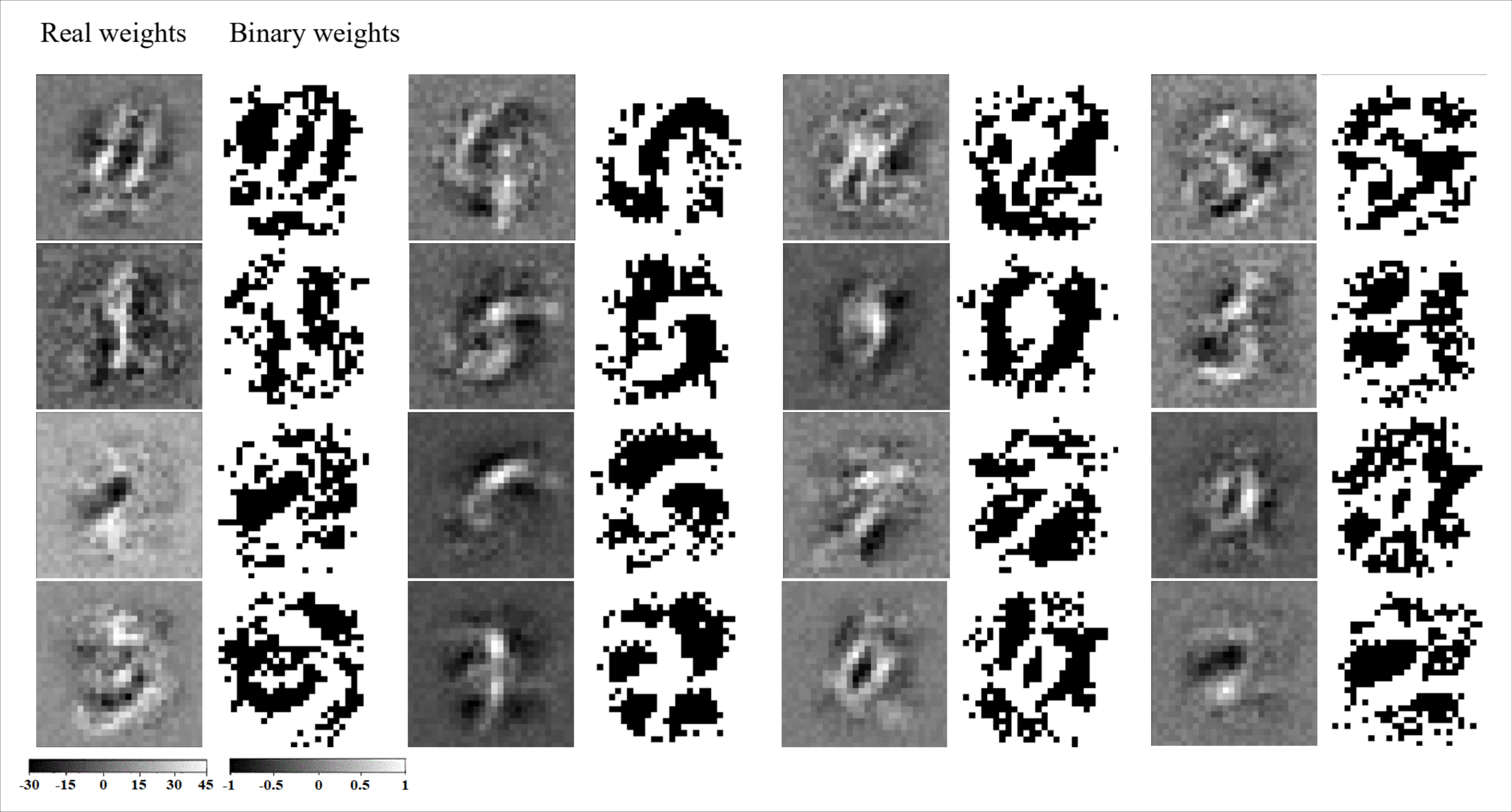}
     \caption{Reconstruction of the real-valued weights and their corresponding binary weights for sixteen randomly selected hidden neurons.}
     \label{fig:5}
 \end{figure}
  \begin{figure}
     \centering
     \includegraphics[width=0.5\textwidth]{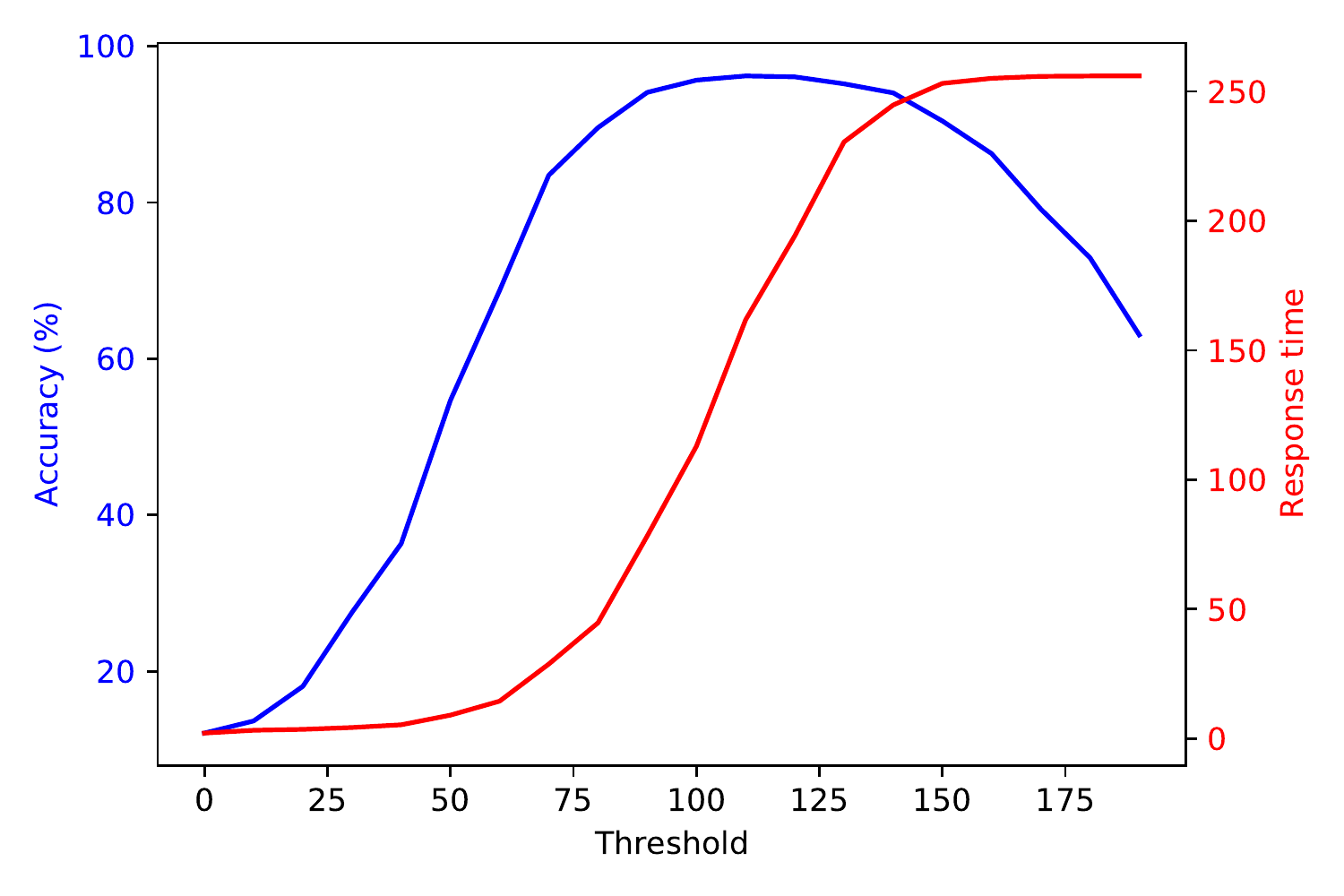}
     \caption{The speed-accuracy trade-off in the pre-trained BS4NN when the threshold varies form 0 to 200.}
     \label{fig:6}
 \end{figure}
 
   \begin{table*}
 \begin{center}
 \caption{The recognition accuracies of recent supervised SNNs on the Fashion-MNIST dataset. The details of each model including its architecture, input coding scheme, neuron model, and learning method are provided. In the ``Hidden Layers" column, letters C, P, R, and F respectively stand for convolutional, pooling, recurrent and fully-connected layers. The number before each layer indicates the number of neurons or convolutional filters in that layer, and the number after the C and P layers indicates their input window size.}\label{tbl2}
 
\resizebox{\textwidth}{!}{ \begin{tabular}{lllllllc}
  &Model&Neuron&Coding&Synapses&Hidden Layers&Learning&  Acc. (\%) \\
 \hline
%Xiao et al. (2017)~\cite{xiao2017fashion}&Fully-connected ANN&ReLU&Decimal &Real-value&Backpropagation&87.1 \\
%%Xiao et al. (2017)~\cite{xiao2017fashion}&Support Vector Machine&-&Supervised&89.7 \\
Zhang et al. (2019)~\cite{zhang2019spike}&Recurrent SNN&Leaky IF&Rate & Real-value&400F-400R&Spike-train backpropagation&90.1\\
Ranjan et al. (2019)~\cite{ranjan2019novel}& Convolutional SNN&Leaky IF&Rate&Real-value&32C3-32C3-P2-32C3-128F& Spike-rate backpropagation&89.0\\
Wu et al. (2020)~\cite{wu2020brain}& Convolutional SNN&Leaky IF& Rate & Real-value&128C3-P2-256C3-
P2-256C3-P2-512F&Global-local hybrid learning rule&93.3\\
Zhang et al. (2020)~\cite{zhang2020temporal}&Fully-connected SNN&Leaky IF&Rate&Real-value&400F-400F&Spike-sequence backpropagation&89.5\\
Zhang et al. (2020)~\cite{zhang2020temporal}&Fully-connected SNN&IOW\footnote{Input Output Weighted Leaky IF~\cite{zhang2020temporal}}&Rate &Real-value&400F-400F&Spike-sequence backpropagation&90.2\\
Hao et al.(2020)~\cite{hao2020biologically}&Fully-connected SNN&Leaky IF&Rate &Real-value&6400F&Dopamine-modulated STDP&85.3\\
S4NN &Fully-connected SNN&IF& Temporal &Real-value&1000F&Temporal backpropagation&88.0\\

BNN&Fully-connected &Binary Sigmoid&Binary&-&1000F&ADAM&86.4\\

BS4NN (this paper)&Fully-connected SNN&IF& Temporal &Binary&1000F&Temporal backpropagation&87.3
 \end{tabular}}
 \end{center}
 \end{table*}

In a further experiment, we replaced the binary weights of the trained BS4NN with their corresponding  real-valued weights and applied them to the test images. In other words, we replaced the $\alpha^{l}B_{ji}^l$ term in Eq.~\ref{Eq:IFNeuronModel} with $W_{ji}^l$. The network reached 89.1\% accuracy on test images which is far less than the 97.0\% accuracy of the binary weights. It shows that, although we update the  real-valued proxy weights during the learning phase, we are actually tuning the binary weights, because the loss and gradients are computed based on the binary weights. Figure~\ref{fig:5} shows the pairs of the real and binary-valued weights for 16 randomly selected hidden neurons. Dark pixels correspond to negative and bright values correspond to positive weights. Here $77.8\%$ of hidden synaptic weights and $60.4\%$ of output synaptic weights are positive.
It seems that hidden neurons tend to detect different variants of digits and their constituent parts.  

To assess the speed-accuracy trade-off in BS4NN, we first trained the network with a threshold of 100 for all the neurons, then we varied the thresholds from 0 to 200 for all the neurons and evaluated the network on test images. As shown in Figure~\ref{fig:6}, the accuracy peaks around the threshold of 100 and drops as we move to higher or lower threshold values, while, the response time (time to the first spike in the output layer) increases by the threshold. Regarding this trade-off, by reducing the threshold of the pre-trained BS4NN, one can get faster responses but with lower accuracy.  For instance, by setting the threshold to 80, the response time shortens from 112.9 to 44.9 ($\sim$3x faster responses), while, the accuracy drops from 97.0\% to 91.0\%.

The $\alpha^l$ scaling factors are full-precision floating-point parameters we used in our neuronal layers to have a better approximation of the real-valued weights by the binary weights. We could round the $\alpha^l$ factors in the pre-trained network down to two decimal places without a  change in the categorization accuracy.
  \begin{figure}
     \centering
     \includegraphics[width=0.48\textwidth]{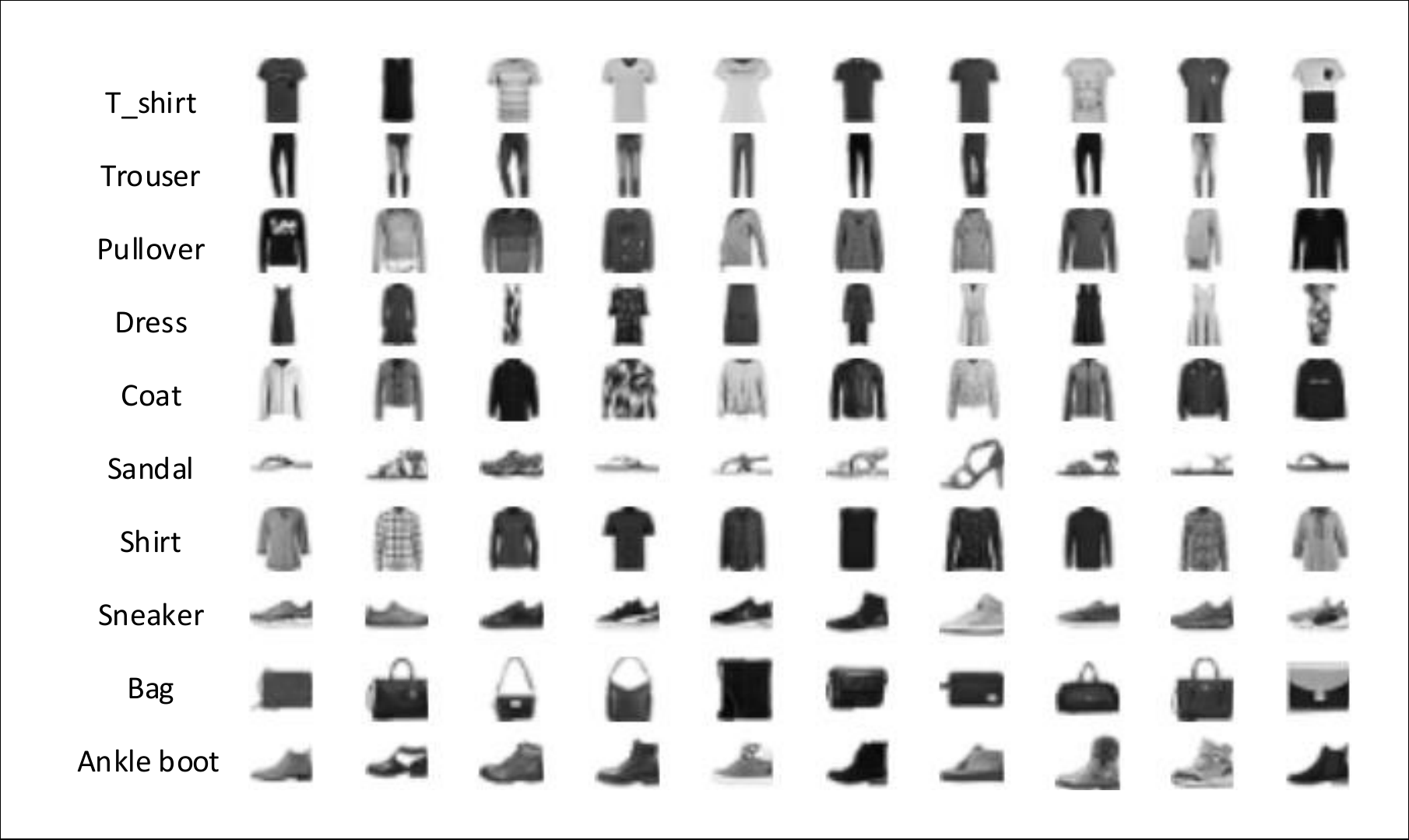}
     \caption{ Sample images from Fashion-MNIST dataset.}
     \label{fig:7}
 \end{figure}

\subsection{Fashion-MNIST dataset}
 Fashion-MNIST~\cite{xiao2017fashion} is a fashion product image dataset with 10 classes (see Figure~\ref{fig:7}). Images are gathered from the thumbnails of the clothing  products on an online shopping website. Fashion-MNIST has the same image size and training/testing splits as MNIST, but it is a more challenging classification task. Here, we used a BS4NN with a single hidden layer with 1000 IF neurons. Details of the parameter values are presented in Table~\ref{table:parameters}. The initial weights of all layers are randomly drawn from a uniform distribution in the range [0,1]. The learning rate parameters $\eta$ and $\mu$ decrease by 30\% every 10 epochs, and the scaling factors $\alpha^{h}$ and $\alpha^o$ are trained during the learning phase.

 \begin{figure}[!h]
     \centering
     \begin{subfigure}{0.5\textwidth}
     \includegraphics[width=\textwidth]{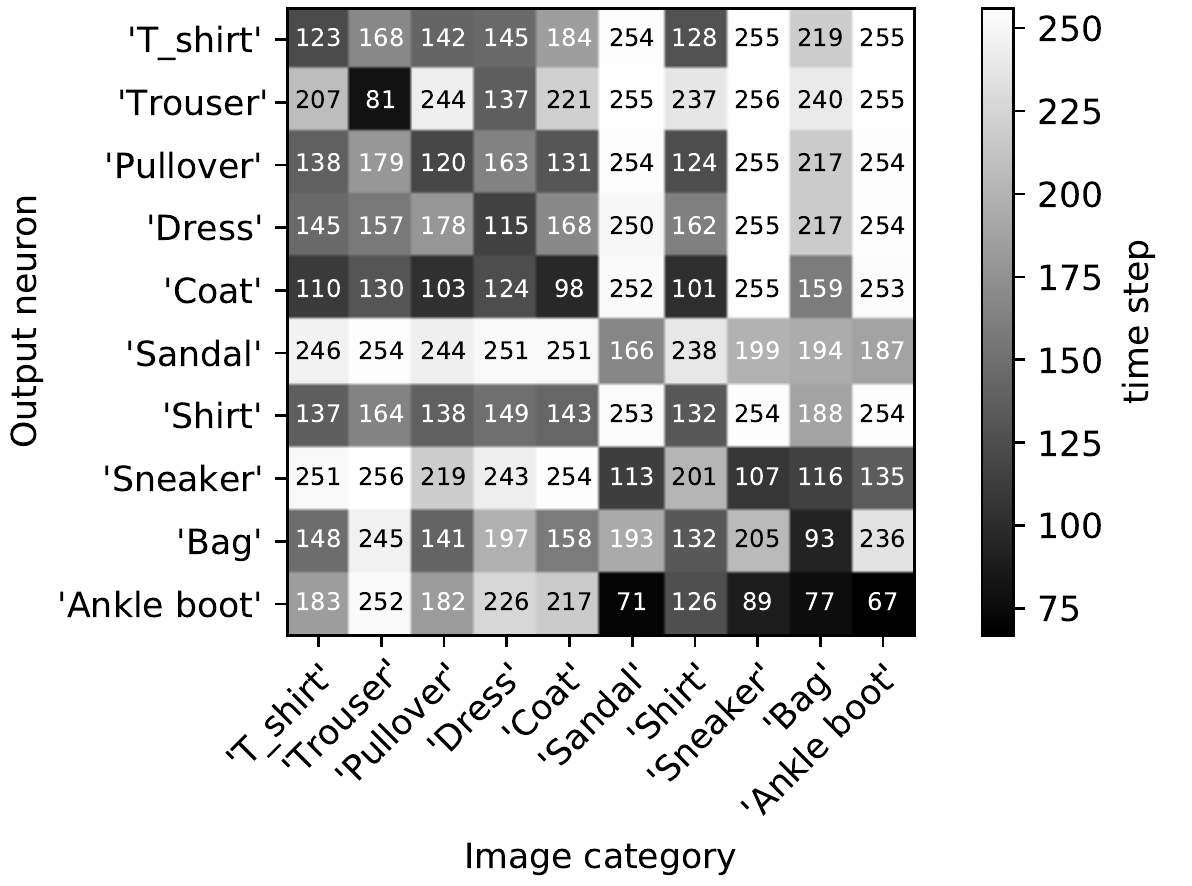}
        \caption{}
        \label{fig:8A}
    \end{subfigure} 
          \begin{subfigure}{0.43\textwidth}
     \includegraphics[width=\textwidth]{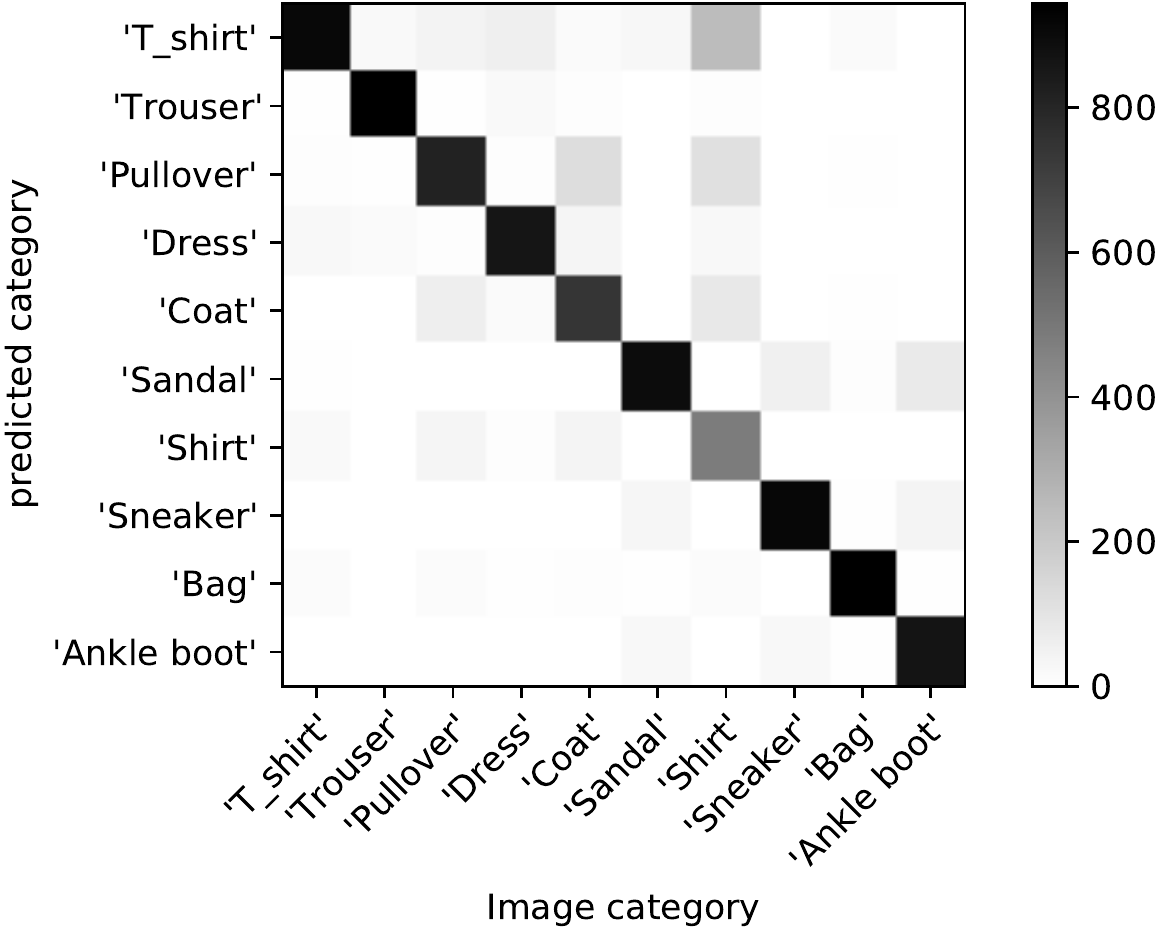}
        \caption{}
        \label{fig:8B}
    \end{subfigure}
     \begin{subfigure}{0.42\textwidth}
     \includegraphics[width=\textwidth]{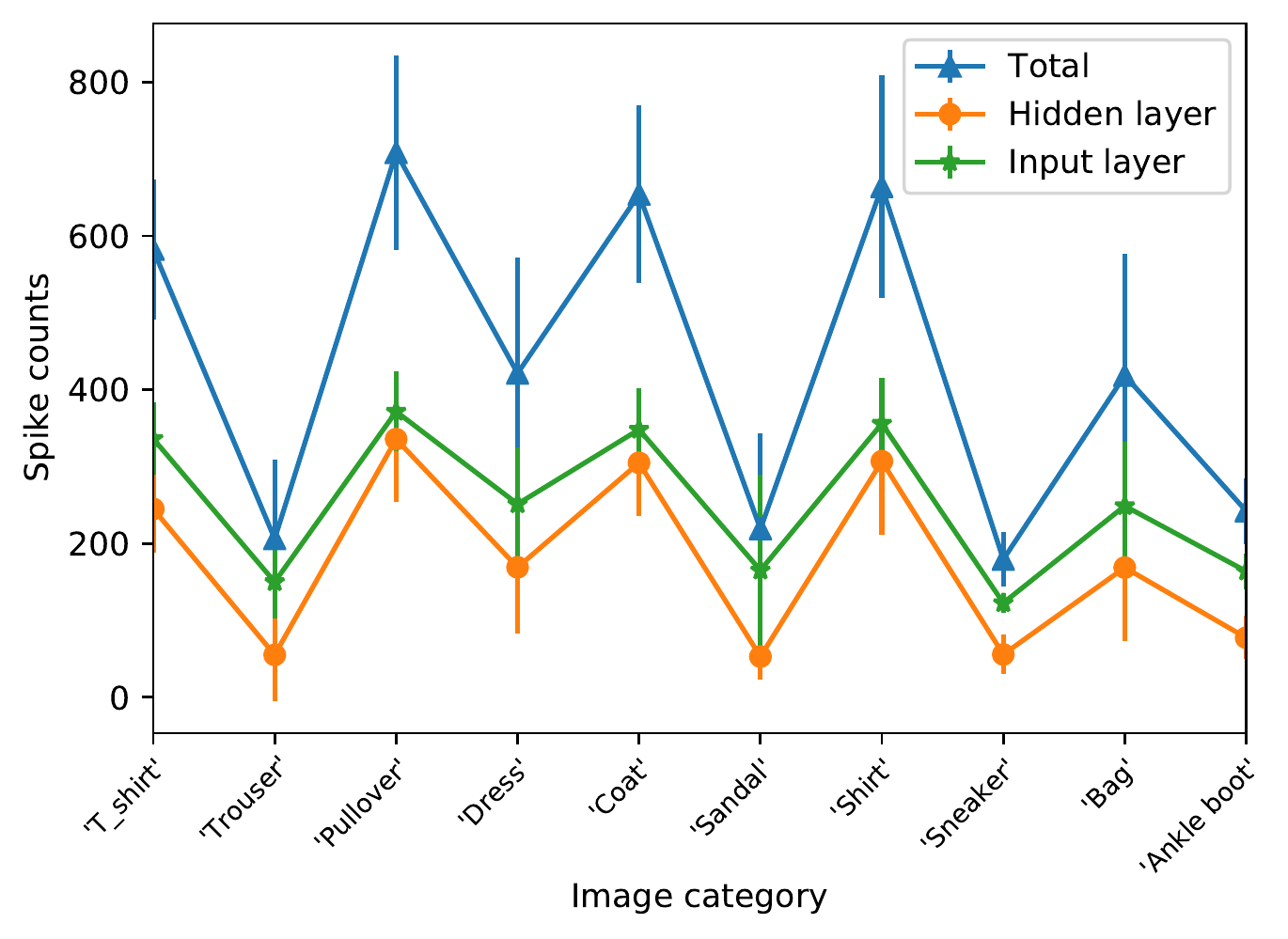}
        \caption{}
        \label{fig:8C}
    \end{subfigure}
\caption{(a) The mean firing times of the output neurons over the Fashion-MNIST categories. (b) The confusion matrix of BS4NN on Fashion-MNIST. (c) The mean required number of spikes per category and layer.}\label{fig:8}
 \end{figure}

 Table~\ref{tbl2}  summarizes the characteristics and recognition accuracies of recent SNNs on the Fashion-MNIST dataset. BS4NN could reach 87.3\% accuracy (0.7\% drop with respect to S4NN). Apart from BS4NN, all the models use real-valued synaptic weights, spike-rate-based neural coding, and leaky neurons with exponential decay.  The mean firing times of the output neurons of BS4NN  for each of the ten categories of Fashion-MNIST are illustrated in Figure~\ref{fig:8A}. As seen, the correct output neuron has the minimum firing time for its corresponding category than others. However, compared to MNIST, there is a small difference between the mean firing times of the correct and some other neurons. It could be due to the similarities between instances of different categories. For instance, as shown in Figure~\ref{fig:8B}, BS4NN confuses ankle boots, sandals, and sneakers. There is a similar situation for shirts and t-shirts, and also, between pullovers and coats, where, their firing times are close together and consequently BS4NN  misclassify them by each other sometimes.  The total required number of spikes in each layer and the total network is provided in Figure~\ref{fig:8C}. Those classes that are mostly confused by each other (i. e., shirts, t-shirts, coats, and pullovers) require more spikes in both input and hidden layers. One reason could be the larger size of these objects in the input image leading to more early input spikes. But the other reason, especially for the hidden layer, could be the need for more discriminative  features between these confusing categories.
 
 We also did a comparison between BS4NN and a BNN with binary weights (-1 and 1), binary activations (0 and1), and the same architecture as BS4NN on Fashion-MNIST.  The BNN is trained using the ADAM optimizer and squared hinge loss function. The learning rate is initially set to $10^{-3}$  and exponential decays down to $10^{-4}$. The initial real-valued weights of each layer are randomly drawn from a uniform distribution in the range  $[-1/\sqrt{n}, 1/\sqrt{n}]$, where, $n$ is the number of synaptic weights of that layer. Interestingly, BS4NN outperforms BNN by 0.9\% accuracy. 

To evaluate the proposed learning algorithm in deeper architectures, we trained and tested a BS4NN network with two fully-connected hidden layers on Fashion-MNIST.
Each hidden layer is consists of $600$ IF neurons with the threshold of $500$.
All the initial weights are randomly drawn
from a uniform distribution in the range [-10; 10] and the scaling factors of each layer are initialized to $10$. The other parameters are the same as the aforementioned two-layer network.
The recognition accuracy on the test set reached $87.5\%$ which shows $0.2\%$ improvement with respect to BS4NN with a single hidden layer (see Table~\ref{tbl2}). It shows that the proposed learning algorithm can well backpropagate the error in deeper architectures.

\section{Discussions}

In this paper, we propose a binarized spiking neural network (called BS4NN) with a direct supervised temporal learning algorithm. To this end, we used a very common approach in the area of BANNs~\cite{simons2019review}. During the learning phase, we have two sets of real and binary-valued weights, such that the binary weights are the sign of the real-valued weights. The binary weights are used for the inference and gradient backpropagation, while, in the backward pass, the weight updates are applied to the real-valued weights.  The proposed BS4NN uses the time-to-first-spike coding~\cite{Mozafari2019a,Mozafari2019b,Vaila2019,vaila2019deep,kirkland2020spikeseg} to convert image pixels into spike trains in which input neurons with higher pixel intensities emit spikes with shorter latencies. The subsequent hidden and output neurons are comprised of non-leaky IF neurons with binary (+1 or -1) weights that fire once when they reach their threshold for the first time. The decision is simply made by the first spike in the output layer.  The temporal error is then computed by comparing the actual and target firing times. Gradients backpropagate through the network and are applied to the real-valued weights. Target firing times are computed relative to the actual firing times of the output neurons to push the correct output neuron to fire earlier than others. It forces BS4NN to make quick and accurate decisions with the less possible amount of spikes (high sparsity). 

In our experiments, BS4NN could reach 97.0\% and 87.3\% accuracy on MNIST and Fashion-MNIST datasets, respectively. Although in terms of accuracy, BS4NN could not beat the real-valued SNNs, it has several computational, memory, and energy advantages which makes it suitable for the inference phase of off-chip trained networks in hardware and neuromorphic implementations. Interestingly, BS4NN has also outperformed BNNs with the same architectures on MNIST and Fashion-MNIST by 0.2\% and 0.9\%  accuracy, respectively. This improvement with respect to BNN could be due to the use of time in our time-to-first-spike coding and temporal backpropagation in BS4NN. Both networks have binary activations and binary weights, but the advantage of BS4NN is the use of temporal information encoded in spike times.

Instead of real-valued weights, BS4NN uses binary synapses with only one full-precision scaling factor per layer. It can be very important for memory optimization in hardware implementations where every synaptic weight requires a separate memory space.  If one implements the binary synapses with a single bit of memory, then it can reduce the network size by 32x compared to a network with 32-bit floating-point weights~\cite{rastegari2016xnor,mcdanel2017embedded}. Also, it can ease the implementation of multiplicative synapses by replacing them with one unit increment and decrement operations. Hence, it can be important for reducing the computational and energy-consumption costs~\cite{rastegari2016xnor,mcdanel2017embedded}.

In BS4NN we used non-leaky IF neurons for three main reasons, 1) we are using temporal coding, and contrary to rate coding, each neuron fires at most once, hence, a few hundred spikes are transmitted throughout the network and more importantly the number of spikes dramatically decreases as we go further in layers and a leak would make this problem even worst, 2) the input stimuli are static images and not videos with dynamic spike trains which a leak could be useful to forget old inputs, and 3) IF neurons are much easier and more energy-efficient to be implemented on hardware (it is already shown that S4NN with IF neurons significantly reduce energy demand on different hardware~\cite{oh2020hardware,liang20211}. The use of non-leaky IF neurons instead of complicated neuron models such as SRM~\cite{comsa2019temporal} and LIF~\cite{zhang2020temporal,masquelier2018optimal} makes BS4NN more computationally efficient and hardware friendly. It might be possible to efficiently implement leakage in analog hardware regarding the physical features of transistors and capacitors~\cite{Roy2019}, but it is always costly to be implemented in digital hardware. To do so, one might periodically (e.g., every millisecond) decrease the membrane potential of all neurons (clock-driven)~\cite{Yousefzadeh2017}, or whenever an input spike is received by a neuron (event-based)~\cite{Orchard2015a,Yousefzadeh2015}. The first one requires energy and the latter one needs more memory to store the last firing times.

The implementation of instantaneous synapses used in BS4NN is way simpler than the exponential~\cite{mostafa2017supervised}, alpha~\cite{comsa2019temporal}, and linear~\cite{zhang2020spike,sakemi2020supervised,rueckauer2018conversion} synaptic currents and costs much less energy and computation. In instantaneous synapses, each input spike causes a sudden potential increment or decrement, but in the current-based synapses, each input spike causes the potential to be updated on several consecutive time steps (which requires an extra state parameter). It is worth mentioning that due to the use of  non-leaky IF neurons with instantaneous synapses, BS4NN does not need any significant adaptation to be implemented in a spike-driven manner or on neuromorphic hardware. Also, as the error is independent of the simulation time in BS4NN, If one doubles the number of time steps, the learning time is not changed. However, other supervised learning rules like surrogate gradient learning backpropagate the error through time and their training time increases with the number of time steps.

As mentioned above, BS4NN uses single-spike neural coding throughout the network. The input layer employs a time-to-first-spike coding by which input neurons fire only once (shorter latencies for stronger inputs). Also, neurons in the subsequent layers are allowed to fire at most once and only when they reach their threshold for the first time. In addition, the proposed temporal learning algorithm used to train BS4NN forces it to rely on earlier spikes and respond as quickly as possible. This cocktail is shown to take much less energy and time on neuromorphic devices compared to the rate-coded SNNs~\cite{chu2020you,goltz2020fast}, even up to 15 times lower energy consumption and  5 times faster decisions~\cite{oh2020hardware}. 

Recently, efforts are made to convert pre-trained BANNs into equivalent BSNNs with spike-rate-based neural coding~\cite{rueckauer2017conversion,wang2020deep,lu2020exploring}. However, these networks do not use the temporal advantages of SNNs that can be obtained through a direct learning algorithm. Due to the non-differentiability of the thresholding activation function in spiking neurons, it is not convenient to apply backpropagation and gradient descents to SNNs. Various solutions are proposed to tackle this problem including computing gradients with respect to the spike rates instead of single spikes~\cite{hunsberger2015spiking,lee2016training,neftci2017event,zenke2018superspike}, using differentiable smoothed spike functions~\cite{huh2018gradient}, surrogate gradients for the threshold function in the backward pass~\cite{Neftci2019,bohte2011error,essera2016convolutional,shrestha2018slayer,bellec2018long,zimmer2019technical,Pellegrini2021,Fang2020b,Zenke2021}, and transfer learning by sharing weights between the SNN and an ANN~\cite{chu2020you,wu2019tandem}. In another approach, known as latency learning, the neuron's activity is defined based on the firing time of its first spike, therefore, they do not need to compute the gradient of the thresholding function. In return, they need to define the firing time as a function of the membrane potential~\cite{Kheradpisheh2020,comsa2019temporal,zhang2020spike,sakemi2020supervised,zhang2020temporal,Bohte2000,zhou2019direct}, or directly as a function of the firing times of presynaptic neurons~\cite{mostafa2017supervised}. All aforementioned learning strategies work with full-precision real-valued weights and future studies can assess their capabilities to be used in BSNNs.


\end{document}